%% file: main.tex
\documentclass[
]{ceurart}
\usepackage{algorithm2e}
\usepackage{diagbox}
\usepackage{url}
\usepackage{adjustbox}
\usepackage{subcaption}
\usepackage{xspace}
\usepackage{import}
\usepackage{paralist}
\usepackage{listings}
\begin{document}

\copyrightyear{2022}
\copyrightclause{Copyright for this paper by its authors.
  Use permitted under Creative Commons License Attribution 4.0
  International (CC BY 4.0).}

\conference{8th Workshop on Practical Aspects of Automated Reasoning}

\newif\ifisPAAR
\isPAARfalse

\ifisPAAR
\title{Exploring Representation of Horn Clauses using GNNs}
\else 
\title{Exploring Representation of Horn Clauses using GNNs (Extended Technical Report)}
\fi

\author[1]{Chencheng Liang}[%
orcid=0000-0002-4926-8089,
email=chencheng.liang@it.uu.se,
url=https://github.com/ChenchengLiang/,
]
\address[1]{Uppsala University, Department of Information Technology, Uppsala, Sweden}
\author[1,2]{Philipp Rümmer}[%
orcid=0000-0002-2733-7098,
email=philipp.ruemmer@it.uu.se,
url=https://github.com/pruemmer/,
]

\address[2]{University of Regensburg, Regensburg, Germany}

\author[3]{Marc Brockschmidt}[%
orcid=0000-0001-6277-2768,
email=marc@marcbrockschmidt.de,
url=https://github.com/mmjb/,
]

\address[3]{Microsoft Research}


\newcommand\hyperedgeHornGraph{control- and data-flow hypergraph\xspace}
\newcommand\hyperedgeHornGraphSpace{control- and data-flow hypergraph }
\newcommand\HyperedgeHornGraph{Control- and data-flow hypergraph\xspace}
\newcommand\HyperedgeHornGraphSpace{Control- and data-flow hypergraph }
\newcommand\HyperedgeHornGraphAbbrev{CDHG\xspace}
\newcommand\HyperedgeHornGraphAbbrevs{CDHGs\xspace}
\newcommand\HyperedgeHornGraphAbbrevSpace{CDHG }
\newcommand\layerHornGraph{constraint graph\xspace}
\newcommand\layerHornGraphs{constraint graphs\xspace}
\newcommand\layerHornGraphSpace{constraint graph }
\newcommand\LayerHornGraph{Constraint graph\xspace}
\newcommand\LayerHornGraphs{Constraint graphs\xspace}
\newcommand\LayerHornGraphSpace{Constraint graph }
\newcommand\layerHornGraphsSpace{constraint graphs }
\newcommand\LayerHornGraphsAbbrev{CG\xspace}

\newcommand\taskOne{Argument identification\xspace}
\newcommand\taskTwo{Count occurrence of relation symbols in all clauses\xspace}
\newcommand\taskThree{Relation symbol occurrence in SCCs\xspace}
\newcommand\taskFour{Existence of argument bounds\xspace}
\newcommand\taskFive{Clause occurrence in counter-examples\xspace}

\newcommand\hyperedgeGNN{R-HyGNN\xspace}
\newcommand\hyperedgeGNNs{R-HyGNNs\xspace}
\newcommand\hyperedgeGNNSapace{R-HyGNN }
\newcommand\hyperedgeGNNsSpace{R-HyGNNs }

\import{sections/}{0-abstract}

\maketitle

\import{sections/}{1-introduction}

\import{sections/}{2-background}
\import{sections/}{3-horn-graphs}

\import{sections/}{4-hyperedge-graph-neural-networks}

\import{sections/}{5-proxy-tasks}

\import{sections/}{6-evaluation}

\import{sections/}{8-related-works}

\import{sections/}{9-conclusion-and-future-works}

\bibliography{main}

\ifisPAAR
\else 
\import{sections/}{abstract-examples-and-algorithms}

\fi

\end{document}

%% file: sections/0-abstract.tex
\begin{abstract}
In recent years, the application of machine learning in program verification, and the embedding of programs to capture semantic information, has been recognised as an important tool by many research groups. Learning program semantics from raw source code is challenging due to the complexity of real-world programming language syntax and due to the difficulty of reconstructing long-distance relational information implicitly represented in programs using identifiers. 
Addressing the first point, we consider Constrained Horn Clauses (CHCs) as a standard representation of program verification problems, providing a simple and programming language-independent syntax.
For the second challenge, we explore graph representations of CHCs, and propose a new Relational Hypergraph Neural Network (R-HyGNN) architecture to learn program features.

We introduce two different graph representations of CHCs. One is called \emph{\layerHornGraph} (\LayerHornGraphsAbbrev), and emphasizes syntactic information of CHCs by translating the symbols and their relations in CHCs as typed nodes and binary edges, respectively, and constructing the constraints as abstract syntax trees. The second one is called \emph{\hyperedgeHornGraphSpace} (\HyperedgeHornGraphAbbrev), and emphasizes semantic information of CHCs by representing the control and data flow through ternary hyperedges.

We then propose a new GNN architecture, \emph{R-HyGNN,} extending Relational Graph Convolutional Networks, to handle hypergraphs. To evaluate the ability of R-HyGNN to extract semantic information from programs, we use \hyperedgeGNNs to train models on the two graph representations, and on five proxy tasks with increasing difficulty, using benchmarks from CHC-COMP~2021 as training data. The most difficult proxy task requires the model to predict the occurrence of clauses in counter-examples, which subsumes satisfiability of CHCs. \HyperedgeHornGraphAbbrev achieves 90.59\% accuracy in this task. Furthermore, R-HyGNN has perfect predictions on one of the graphs consisting of more than 290 clauses. Overall, our experiments indicate that R-HyGNN can capture intricate program features for guiding verification problems.

\end{abstract}

\begin{keywords}
  Constraint Horn clauses \sep
  Graph Neural Networks \sep
  Automatic program verification
\end{keywords}

%% file: sections/1-introduction.tex
\section{Introduction}

Automatic program verification is challenging because of the complexity of industrially relevant programs. In practice, constructing domain-specific heuristics from program features (e.g., information from loops, control flow, or data flow) is essential for solving verification problems. For instance, \cite{Leroux2016} and~\cite{10.1007/978-3-319-57288-8_18} extract semantic information by performing systematical static analysis to refine abstractions for the counterexample-guided abstraction refinement (CEGAR)~\cite{10.1007/10722167_15} based system.
However, manually designed heuristics usually aim at a specific domain and are hard to transfer to other problems. Along with the rapid development of deep learning in recent years, learning-based methods have evolved quickly and attracted more attention. For example, the program features are explicitly given in \cite{10.1145/2597073.2597080, mci/Demyanova2016} to decide which algorithm is potentially the best for verifying the programs. Later in \cite{Richter2020-yh,9286080}, program features are learned in the end-to-end pipeline. Moreover, some generative models~\cite{https://doi.org/10.48550/arxiv.1312.6114, https://doi.org/10.48550/arxiv.1802.08786} are also introduced to produce essential information for solving verification problems. For instance,  
Code2inv~\cite{CodeToInv} embeds the programs by graph neural networks (GNNs) \cite{DBLP:journals/corr/abs-1806-01261} and learns to construct loop invariants by a deep neural reinforcement framework.

For deep learning-based methods, no matter how the learning pipeline is designed and the neural network structure is constructed, learning to represent semantic program features is essential and challenging
 (a) because the syntax of the source code varies depending on the programming languages, conventions, regulations, and even syntax sugar and
 (b) because it requires capturing intricate semantics from long-distance relational information based on re-occurring identifiers.
For the first challenge, since the source code is not the only way to represent a program, learning from other formats is a promising direction. For example, inst2vec~\cite{DBLP:journals/corr/abs-1806-07336} learns control and data flow from LLVM intermediate representation~\cite{1281665} by recursive neural networks (RNNs)~\cite{Mikolov2010RecurrentNN}.
Constrained Horn Clauses (CHCs)~\cite{10.2307/2268661}, as an intermediate verification language, consist of logic implications and constraints and can alleviate the difficulty since they can naturally encode program logic with simple syntax. For the second challenge, we use graphs to represent CHCs and learn the program features by GNNs since they can learn from the structural information within the node's N-hop neighbourhood by recursive neighbourhood aggregation (i.e., neural message passing) procedure.

In this work, we explore how to learn program features from CHCs by answering two questions: (1) What kind of graph representation is suitable for CHCs? (2) Which kind of GNN is suitable to learn from the graph representation?

For the first point, we introduce two graph representations for CHCs: the \layerHornGraph (\LayerHornGraphsAbbrev) and \hyperedgeHornGraphSpace (\HyperedgeHornGraphAbbrev). The \layerHornGraph encodes the CHCs into three abstract layers (predicate, clause, and constraint layers) to preserve as much structural information as possible (i.e., it emphasizes program syntax). On the other hand, the \HyperedgeHornGraph uses ternary hyperedges to capture the flow of control and data in CHCs to emphasize program semantics. To better express control and data flow in \HyperedgeHornGraphAbbrev, we construct it from normalized CHCs. The normalization changes the format of the original CHC but retains logical meaning. 
We assume that different graph representations of CHCs capture different aspects of semantics. The two graph representations can be used as a baseline to construct new graph representations of CHC to represent different semantics. In addition, similar to the idea in~\cite{48724}, our graph representations are invariant to the concrete symbol names in the CHCs since we map them to typed nodes.

For the second point, we propose a Relational Hypergraph Neural Network (\hyperedgeGNN), an extension of Relational Graph Convolutional Networks (R-GCN)~\cite{schlichtkrull2017modeling}. Similar to the GNNs used in LambdaNet~\cite{wei20206lambdanet}, \hyperedgeGNN can handle hypergraphs by concatenating the node representations involved in a hyperedge and passing the messages to all nodes connected by the hyperedge.

Finally, we evaluate our framework (two graph representations of CHCs and \hyperedgeGNN) by five proxy tasks (see details in Table \ref{tab:five-tasks}) with increasing difficulties. Task~1 requires the framework to learn to classify syntactic information of CHCs, which is explicitly encoded in the two graph representations. Task~2 requires the \hyperedgeGNN to predict a syntactic counting task. Task~3 needs the \hyperedgeGNN to approximate the Tarjan's algorithm~\cite{Tarjan1972DepthFirstSA}, which solves a general graph theoretical problem. Task~4 is much harder than the last three tasks since the existence of argument bounds is undecidable. Task~5 is harder than solving CHCs since it predicts the trace of counter-examples (CEs). Note that Task 1 to 3 can be easily solved by specific, dedicated standard algorithms. We include them
to systematically study the representational power of graph neural networks applied to different graph construction methods.
However, we speculate that using these tasks as pre-training objectives for neural networks that are later fine-tuned to specific (data-poor) tasks may be a successful strategy which we plan to study in future work.

\begin{table*}[tb]
\caption{Proxy tasks used to evaluate suitability of different graph representations.}
\label{tab:five-tasks}
\begin{tabular}{@{}p{4cm}p{2cm}p{7cm}@{}}
\hline
Task & Task type & Description\\
\hline
1.~\taskOne  &  Node binary classification & For each element in CHCs, predict if it is an argument of relation symbols.  \\
\hline
2.~\taskTwo  &   Regression task on node & For each relation symbol, predict how many times it occurs in all clauses. \\
\hline
3.~\taskThree &   Node binary classification  &    For each relation symbol, predict if a cycle exists from the node to itself (membership in strongly connected component, SCC).  \\
\hline
4.~\taskFour &   Node binary classification   & For each argument of a relation symbol, predict if it has a lower or upper bound. \\
\hline
5.~\taskFive  &   Node binary classification  & For each CHC, predict if it appears in counter-examples.  \\
\hline
\end{tabular}
\end{table*}

Our benchmark data is extracted from the 8705~linear and 8425~non-linear Linear Integer Arithmetic (LIA) problems in the CHC-COMP repository\footnote{\url{https://chc-comp.github.io/}} (see Table~1 in the competition report~\cite{chcBenchmark}). The verification problems come from various sources (e.g., higher-order program verification benchmark\footnote{\url{https://github.com/chc-comp/hopv}} and benchmarks generated with JayHorn\footnote{\url{https://github.com/chc-comp/jayhorn-benchmarks}}), therefore cover programs with different size and complexity. 
We collect and form the train, valid, and test data using the predicate abstraction-based model checker Eldarica~\cite{ruemmer2013disjunctive}. We implement \hyperedgeGNNs\footnote{\url{https://github.com/ChenchengLiang/tf2-gnn}} based on the framework tf2\_gnn\footnote{\url{https://github.com/microsoft/tf2-gnn}}. Our code is available in a Github repository\footnote{\url{https://github.com/ChenchengLiang/Systematic-Predicate-Abstraction-using-Machine-Learning}}.
For both graph representations, even if the predicted accuracy decreases along with the increasing difficulty of tasks, 
for undecidable problems in Task 4, \hyperedgeGNN still achieves high accuracy, i.e., 91\% and 94\% for \layerHornGraph and \HyperedgeHornGraphAbbrev, respectively. Moreover, in Task 5, despite the high accuracy (96\%) achieved by \HyperedgeHornGraphAbbrev, \hyperedgeGNN has a perfect prediction on one of the graphs consisting of more than 290 clauses, which is impossible to achieve by learning simple patterns (e.g., predict the clause including $\textit{false}$ as positive). Overall, our experiments show that our framework learns not only the explicit syntax but also intricate semantics.

\paragraph{Contributions of the paper.}
\begin{inparaenum}[(i)]
  \item We encode CHCs into two graph representations, emphasising abstract program syntactic and semantic information, respectively.
  \item We extend a message passing-based GNN, R-GCN, to \hyperedgeGNN to handle hypergraphs.
  \item We introduce five proxy supervised learning tasks to explore the capacity of \hyperedgeGNN to learn semantic information from the two graph representations.
  \item We evaluate our framework on the CHC-COMP benchmark and show that this framework can learn intricate semantic information from CHCs and has the potential to produce good heuristics for program verification.
\end{inparaenum}

%% file: sections/2-background.tex
\section{Background} \label{section-banckground}

\subsection{From Program Verification to Horn clauses}
Constrained Horn Clauses are logical implications involving unknown predicates. They can be used to encode many formalisms, such as transition systems, concurrent systems, and interactive systems. The connections between program logic and CHCs can be bridged by Floyd-Hoare logic~\cite{Floyd1967Flowcharts,10.1145/363235.363259}, allowing to encode program verification problems into the CHC satisfiability problems~\cite{Bjorner2015}. In this setting, a program is guaranteed to satisfy a specification if the encoded CHCs are satisfiable, and vice versa.

We write CHCs in the form
$H \leftarrow B_{1}  \wedge \cdots \wedge B_{n} \wedge \varphi$,
where
\begin{inparaenum}[(i)]
  \item $B_{i}$ is an application $q_{i}(\bar{t_{i}})$ of the relation symbol $q_{i}$ to a list of first-order terms $\bar{t_{i}}$;
  \item $H$ is either an application $q(\bar{t})$, or $\mathit{false}$;
  \item $\varphi$ is a first-order constraint.
\end{inparaenum}
Here, $H$ and $B_{1} \wedge \cdots \wedge B_{n} \wedge \varphi$ in the left and right hand side of implication $\leftarrow$ are called ``head" and ``body", respectively.

An example in Figure~\ref{from-program-to-Horn} explains how to encode a verification problem into CHCs.
In Figure~\ref{while-C-program}, we have a verification problem, i.e., a C program with specifications. It has an external input $n$, and we can initially assume that $x==n, y==n, \text{and, } n>=0$. While $x$ is not equal to 0, $x$ and $y$ are decreased by 1. The assertion is that finally, $y==0$.
This program can be encoded to the CHC shown in Figure~\ref{while-C-program-Horn-clauses}. The variables $x \text{ and } y$ are quantified universally.
We can further simplify the CHCs in Figure~\ref{while-C-program-Horn-clauses} to the CHCs shown in Figure~\ref{while-C-program-Horn-clauses-simplified} without changing the satisfiability by some preprocessing steps (e.g., inlining and slicing)~\cite{8603013}. For example, the first CHC encodes line 3, i.e., the assume statement, the second clause encodes lines 4-7, i.e., the while loop, and the third clause encodes line 8, i.e., the assert statement in Figure~\ref{while-C-program}. Solving the CHCs is equivalent to answering the verification problem. In this example, with a given $n$, if the CHCs are satisfiable for all $x \text{ and } y$, then the program is guaranteed to satisfy the specifications.

\begin{figure*}[tb]
\begin{adjustbox}{minipage=0.485\textwidth,frame}
\begin{subfigure}{\textwidth}
\begin{lstlisting}
extern int n;
void main(){
    int x,y;
    assume(x==n && y==n && n>=0);
    while(x!=0){
        x--;
        y--;
    }
    assert(y==0);
}
\end{lstlisting}\subcaption{An verification problem written in C.}\label{while-C-program}
\end{subfigure}
\end{adjustbox}
\begin{adjustbox}{minipage=0.5\textwidth,frame}
\begin{subfigure}{1\textwidth}
\vspace*{-2ex}
\begin{eqnarray*}\label{grammar}
\begin{aligned}
 L_{0}(n)  &\leftarrow  true & \text{\color{gray}line 0} &\\
 L_{1}(n)  &\leftarrow  L_{0}(n) & \text{\color{gray}line 1}\\
 L_{2}(x, y, n)  &\leftarrow  L_{1}(n) & \text{\color{gray}line 2}\\
 L_{3}(x, y, n)  &\leftarrow  L_{2}(x,y,n) \wedge n \geq 0 \\
 & ~~~~~ \wedge x = n \wedge y = n & \text{\color{gray}line 3} \\
 L_{8}(x, y, n)  &\leftarrow  L_{3}(x, y, n) \wedge x = 0  & \text{\color{gray}line 4}\\
 L_{4}(x, y, n)  &\leftarrow  L_{3}(x, y, n) \wedge x \neq 0  & \text{\color{gray}line 4}\\
 L_{5}(x, y, n)  &\leftarrow  L_{4}(x', y, n) \wedge x = x'-1  & \text{\color{gray}line 5}\\
 L_{6}(x, y, n)  &\leftarrow  L_{5}(x, y', n) \wedge y = y'-1  & \text{\color{gray}line 6}\\
 L_{3}(x, y, n)  &\leftarrow  L_{6}(x, y, n)  & \text{\color{gray}line 6}\\
 \mathit{false}  & \leftarrow  L_{8}(x, y, n) \wedge y \neq 0 & \text{\color{gray}line 8}\\
\end{aligned}
\end{eqnarray*}
\vspace*{-1ex}
\subcaption{CHCs encoded from C program in Figure~\ref{while-C-program}. }\label{while-C-program-Horn-clauses}
\end{subfigure}
\end{adjustbox}

\begin{subfigure}{1\textwidth}
\begin{adjustbox}{minipage=1.0\textwidth,frame}
\vspace*{-2ex}
\begin{eqnarray*}\label{grammar}
\begin{aligned}
 L(x,y,n)  & \leftarrow  n \geq 0 \wedge x = n \wedge y = n & \text{\color{gray}line 3~~~} \\
 L(x,y,n)  & \leftarrow  L(x',y',n') \wedge x' \neq 0 \wedge x=x' -1 \wedge y=y' -1 \wedge n=n' & \text{\color{gray}line 4-7} \\
 \mathit{false}  & \leftarrow  L(x,y,n) \wedge x=0 \wedge y \neq 0 & \text{\color{gray}line 8~~~} \\
\end{aligned}
\end{eqnarray*}
\vspace*{-1ex}
\subcaption{Simplified CHCs from  Figure~\ref{while-C-program-Horn-clauses}.}\label{while-C-program-Horn-clauses-simplified}
\end{adjustbox}
\end{subfigure}
\caption{An example to show how to encode a verification problem written in C to CHCs. For the C program, the left-hand side numbers indicate the line number. The line numbers in Figure~\ref{while-C-program-Horn-clauses} and ~\ref{while-C-program-Horn-clauses-simplified} correspond to the line in Figure~\ref{while-C-program}. For example, the line $L_{0}(n) \leftarrow  true$ in Figure~\ref{while-C-program-Horn-clauses} is transformed from line 1 ``extern int n ;" in Figure~\ref{while-C-program}. } \label{from-program-to-Horn}
\end{figure*}

\subsection{Graph Neural Networks}
Let $G=(V,R,E,X,\ell)$ denote a graph in which 
 $v\in V$ is a set of nodes,
 $r\in R$ is a set of edge types,
 $E\in V \times V \times R$ is a set of typed edges,
 $x\in X$ is a set of node types, and 
 $\ell: v\rightarrow x$ is a labelling map from nodes to their type.
A tuple $e=(u,v,r)\in E$ denotes an edge from node $u$ to $v$ with edge type $r$.

Message passing-based GNNs use a neighbourhood aggregation strategy, where at timestep $t$, each node updates its representation $h_v^t$ by aggregating representations of its neighbours and then combining its own representation.
The initial node representation $h_v^0$ is usually derived from its type or label $\ell(v)$.
The common assumption of this architecture is that after $T$ iterations, the node representation $h_{v}^{T}$ captures local information within $t$-hop neighbourhoods.
Most GNN architectures~\cite{DBLP:journals/corr/GilmerSRVD17,DBLP:journals/corr/abs-1810-00826} can be characterized by their used ``aggregation'' function $\rho$ and ``update'' function $\phi$.
The node representation of the $t$-th layer of such a GNN is then computed by $h_{v}^{t} = \phi(\rho(\{h_{u}^{t-1}\mid u \in N_{v}^{r},{r\in R}\}), h_{v}^{t-1})$,
where $R$ is a set of edge type and $N_{v}^{r}$ is the set of nodes that are the neighbors of $v$ in edge type $r$.

A closed GNN architecture to the \hyperedgeGNN is R-GCN~\cite{schlichtkrull2017modeling}.
They update the node representation by
\begin{equation}\label{R-GCN-GNN}
  h_{v}^{t} = \sigma(\sum_{r\in R}\sum_{u \in N_{v}^{r}} \frac{1}{c_{v,r}}W_{r}^{t}h_{u}^{t-1}+W_{0}h_{v}^{t-1}),
\end{equation}
where $W_{r}$ and $W_{0}$ are edge-type-dependent trainable weights, $c_{v,r}$ is a learnable or fixed normalisation factor, and $\sigma$ is a activation function. 

%% file: sections/3-horn-graphs.tex
\newcommand{\icon}[2][0.5cm]{\begin{minipage}{#1}\includegraphics[width=#1]{#2}\end{minipage}}

\section{Graph Representations for CHCs}\label{sec:graphs}
Graphs as a representation format support arbitrary relational structure and thus can naturally encode information with rich structures like CHCs.
We define two graph representations for CHCs that emphasize the program syntax and semantics, respectively. We map all symbols in CHCs to typed nodes and use typed edges to represent their relations.
In this section, we give concrete examples to illustrate how to construct the two graph representations from a single CHC modified from Figure \ref{while-C-program-Horn-clauses-simplified}. In the examples, we first give the intuition of the graph design and then describe how to construct the graph step-wise. To better visualize how to construct the two graph representations in Figures~\ref{layer-graph-step-example} and \ref{hyperedge-graph-step-example}, the concrete symbol names for the typed nodes are shown in the blue boxes. \hyperedgeGNN is not using these names (which, as a result of repeated transformations, usually do not carry any meaning anyway) and only consumes the node types.
\ifisPAAR
  The formal definitions of the graph representations and the algorithms to construct them from multiple CHCs are in the full version of this paper \cite{tech-report}.
\else 
  We include abstract examples, the formal definitions of the graph representations, and the algorithms to construct them from multiple CHCs in this section as well.
\fi
Note that the two graph representations in this study are designed empirically. They can be used as a baseline to create variations of the graphs to fit different purposes.

\subsection{Constraint Graph (CG)}
\begin{figure*}[tb]
\centering
  \includegraphics[width=\linewidth]{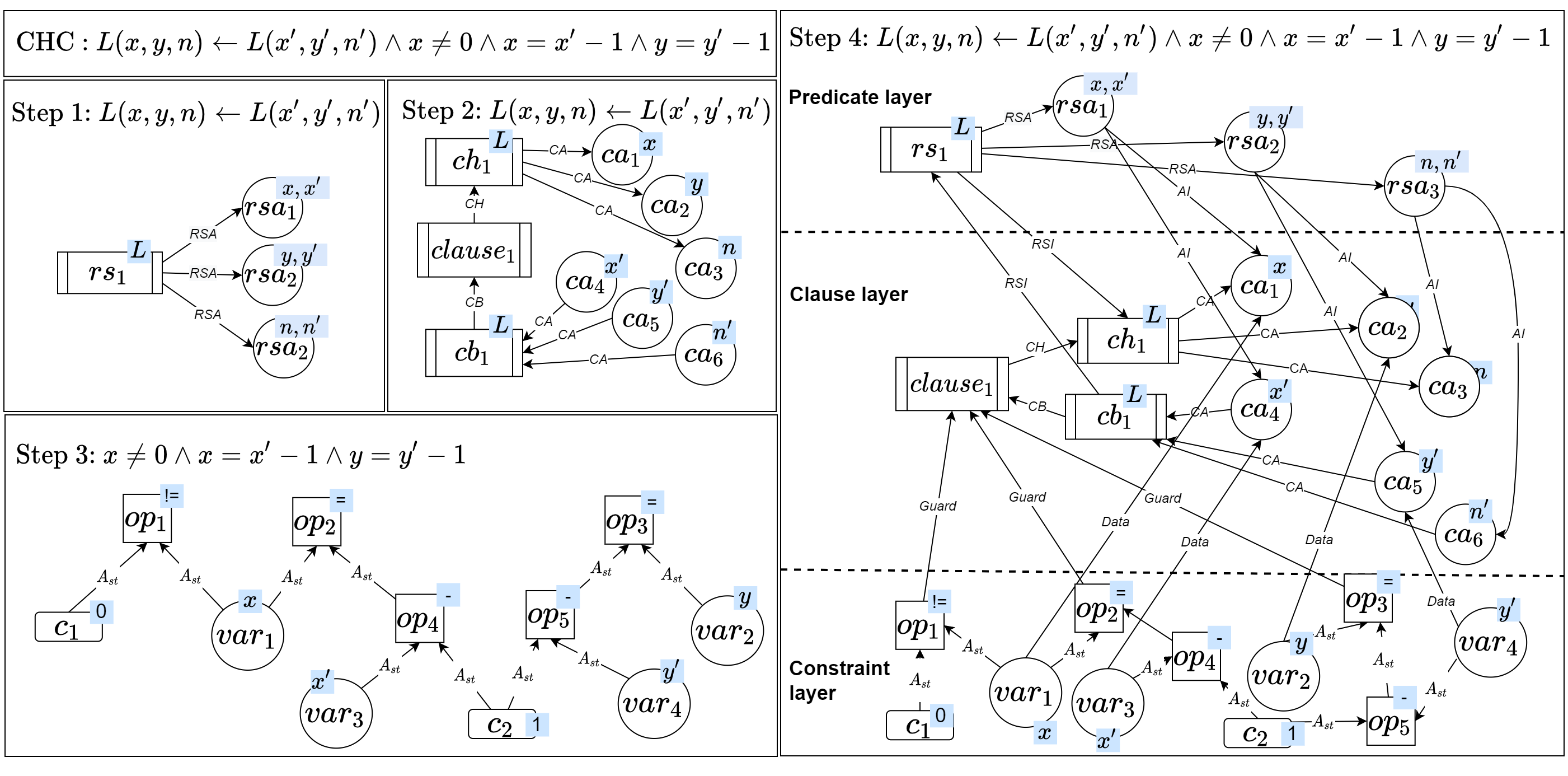}\\
  \caption{Construct \layerHornGraph from the CHC $L(x,y,n)  \leftarrow  L(x',y',n') \wedge x \neq 0 \wedge x=x' -1 \wedge y=y' -1$. Note that some nodes have multiple concrete symbol names (e.g., node $rsa_{1}$ has two concrete names, $x$ and $x'$) since one relation symbol may bind with different arguments.}\label{layer-graph-step-example}
\end{figure*}

Our \LayerHornGraph is a directed graph with binary edges designed to emphasize syntactic information in CHCs.
One concrete example of constructing the \layerHornGraph for a single CHC $L(x,y,n)  \leftarrow  L(x',y',n') \wedge x \neq 0 \wedge x=x' -1 \wedge y=y' -1$ modified from Figure~\ref{while-C-program-Horn-clauses-simplified} is shown in Figure~\ref{layer-graph-step-example}.
\ifisPAAR
  The corresponding node and edge types are described in Tables~2 and~3 in the full version of this paper \cite{tech-report}.
\else 
  The corresponding node and edge types are described in Tables~\ref{layer-Horn-graph-node-definition} and~\ref{layer-Horn-graph-edge-definition}.
\fi

We construct the \layerHornGraph by parsing the CHCs in three different aspects (relation symbol, clause structure, and constraint) and building relations for them.
In other words, a \layerHornGraph consists of three layers: the predicate layer depicts the relation between relation symbols and their arguments; the clause layer describes the abstract syntax of head and body items in the CHC; the constraint layer represents the constraint by abstract syntax trees (ASTs).

\paragraph{Constructing a \layerHornGraph.}
Now we give a concrete example to describe how to construct a \layerHornGraph for a single CHC $L(x,y,n)  \leftarrow  L(x',y',n') \wedge x \neq 0 \wedge x=x' -1 \wedge y=y' -1$ step-wise. All steps correspond to the steps in Figure~\ref{layer-graph-step-example}.
In the first step, we draw relation symbols and their arguments as typed nodes and build the connection between them.
In the second step, we construct the clause layer by drawing clauses, the relation symbols in the head and body, and their arguments as typed nodes and build the relation between them.
In the third step, we construct the constraint layer by drawing ASTs for the sub-expressions of the constraint.
In the fourth step, we add connections between three layers. The predicate and clause layer are connected by the relation symbol instance ($\textit{RSI}$) and argument instance ($\textit{AI}$) edges, which means the elements in the predicate layer are instances of the clause layer.
The clause and constraint layers are connected by the $\textit{GUARD}$ and $\textit{DATA}$ edges since the constraint is the guard of the clause implication, and the constraint and clause layer share some elements.

\ifisPAAR

\else 

\paragraph{Formal definition of \layerHornGraphSpace.}
A \layerHornGraphSpace $CG=(V,\mathit{BE},R^{CG},X^{CG},\ell)$ consists of a set of nodes $v\in V$, a set of typed binary edges  $\mathit{BE} \in V \times V \times R^{CG}$, a set of edge types $r \in R^{CG}$ (Table~\ref{layer-Horn-graph-edge-definition}), a set of node types $x\in X^{CG}$ (Table~\ref{layer-Horn-graph-node-definition}), and a map $\ell: v \rightarrow x$.
Here, $(v_{1},v_{2},r) \in \mathit{BE}$ denotes a binary edge with edge type $r$. The node types are used to generate the initial feature $x_{v}$, a real-valued vector in \hyperedgeGNN.

\begin{table*}[tb]
\caption{Node types for \layerHornGraph. The shape corresponds to the shape of nodes in Figure~\ref{layer-graph-step-example} and is only used for visualizing the example.} \label{layer-Horn-graph-node-definition}
\begin{tabular}{p{2.5cm}lp{4cm}lp{1cm}}
\hline
Node types $X^{\LayerHornGraphsAbbrev}$ & Layer & Explanation &  Elements in CHCs & Shape \\
\hline
 \textit{relation symbol} ($rs$) & Predicate layer & Relation symbols in head or body  &  $L$   & \icon{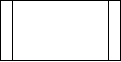}   \\
 \textit{false}  & Predicate layer &  \textit{false} state  &   \textit{false}       & \icon{graph/process}     \\
 \textit{relation symbol argument} ($rsa$) & Predicate layer  & Arguments of the relation symbols &   x, y   & \icon{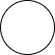}  \\
 \textit{clause} ($cla$) & Clause layer &  Represent clause as a abstract node  & $\emptyset$       & \icon{graph/process}     \\
 \textit{clause head} ($ch$) & Clause layer &  Relation symbol in head  &  $L$          & \icon{graph/process}  \\
 \textit{clause body} ($cb$) & Clause layer &  Relation symbol in body  &  $L$         & \icon{graph/process}   \\
 \textit{clause argument} ($ca$)  & Clause layer &  Arguments of relation symbol in head and body  &   $x,y$      & \icon{graph/circle}      \\
 \textit{variable} ($var$) & Constraint layer & Free variables  &  n   & \icon{graph/circle}  \\
 \textit{operator} ($op$) & Constraint layer  &  Operators over a theory &   =, -      & \icon{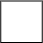}   \\
 \textit{constant} ($c$)  & Constraint layer &   Constants over a theory &  0, 1, $\textit{true}$   & \icon{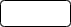} \\
\hline
\end{tabular}
\end{table*}

\begin{table*}[tb]
\caption{Edge types for \layerHornGraph. Here, $rs,rsa,cla$, etc. are node types described in Table~\ref{layer-Horn-graph-node-definition}.Here, ``$|$" means ``or". For example, $c~|~op~|var$ means this node could be node with type $c$, $op$, or $var$.}
\label{layer-Horn-graph-edge-definition}
\begin{tabular}{p{2.8cm}p{2.8cm}p{3.5cm}p{4cm}}
\hline
Edge type $R^{\LayerHornGraphsAbbrev}$ & Layer & Definition & Explanation \\
\hline
Relation Symbol Argument (\textit{RSA})  & Predicate layer & ($rs,rsa, \textit{RSA}$)  &  Connects relation symbols and their arguments \\
\hline
Relation Symbol Instance (\textit{RSI})  & Between predicate and clause layer & $(rs,ch,\textit{RSA})~|~(cb,rs,\textit{RSA})$  & Connects relation symbols with their head and body\\
\hline
Argument~Instance (\textit{AI})  &  Between predicate and clause layer & ($pa,ca,\textit{AI}$)  &  Connects relation symbols and their arguments \\
\hline
Clause~Head (\textit{CH})  &  Clause layer & $(cla,ch,\textit{CH})$  &  Connect $clause$ node to its head  \\
\hline
Clause~Body (\textit{CB})  &  Clause layer & $(cb,cla,\textit{CB})$  &  Connect the $clause$ node to its body  \\
\hline
Clause Argument (\textit{CA})  & Clause layer & $(ca,cb,\textit{CA})~|~(ch,ca,\textit{CA})$ &   Connect $ch$ or $cb$ nodes with corresponding $ca$ nodes \\
\hline
Guard (\textit{GUARD})  &  Between clause and constraint layer & $(c~|~op,cla,\textit{GUARD})$ &  Connect the root node of the AST to corresponding $clause$ node \\
\hline
Data (\textit{DATA})  &  Between clause and constraint layer & $(var,ca,\textit{DATA})$  &  Connect $ca$ nodes to corresponding $var$ nodes AST\\
\hline
AST~sub-term ($A_{st}$) & Constraint layer & $(c~|~var~|~op,op,A_{st})$  & Connect nodes within AST  \\
\hline
\end{tabular}
\end{table*}

\paragraph{An abstract example of \layerHornGraphSpace.}
Except for the concrete example, we give a abstract example to describe how to construct the \layerHornGraph.
First, the CHC $H \leftarrow B_{1}  \wedge \cdots \wedge B_{n} \wedge \varphi$ in Section~\ref{section-banckground} can be re-written to
\begin{equation}\label{horn-clause-definition-3}
 q_{1}(\bar{t_{1}}) \leftarrow q_{2}(\bar{t_{2}}) \wedge \cdots \wedge q_{k}(\bar{t_{k}}) \wedge \varphi_{1}  \wedge \cdots \varphi_{n}, (n,k \in \mathbb{N}),
\end{equation}
where $\varphi_{1} \cdots \varphi_{k}$ are sub-formulas for constraint $\varphi$. Notice that since there is no normalization for the original CHCs, the same relation symbols can appear in both head and body (i.e., $q_{1}, q_{2}, \cdots, q_{k}$ may equal to each other).

Then, we can construct a \layerHornGraph using Algorithm~\ref{layer-graph-algorithm}, in which the input is a set of CHC and the output is a \layerHornGraph $CG=(V,\mathit{BE},R^{CG},X^{CG},\ell)$.
The step-wise constructing process for the CHC in Eq.~(\ref{horn-clause-definition-3}) is visualized in Figure~\ref{abstract-layer-horn-graph}.
\fi

\subsection{Control- and Data-flow Hypergraph (\HyperedgeHornGraphAbbrev)}
In contrast to the \layerHornGraph, the \HyperedgeHornGraphAbbrev representation emphasizes the semantic information (control and data flow) in the CHCs by hyperedges which can join any number of vertices.
%
%
To represent control and data flow in \HyperedgeHornGraphAbbrev, first, we preprocess every CHC and then split the constraint into control and data flow sub-formulas. 

\paragraph{Normalization.}

We normalize every CHC by applying the following rewriting steps:
\begin{inparaenum}[(i)]
  \item We ensure that every relation symbol occurs at most once in every clause. For instance, the CHC $q(a)\leftarrow q(b)\wedge q(c)$ has multiple occurrences of the relation symbol $q$, and we normalize it to equi-satisfiable CHCs $q(a)\leftarrow q'(b)\wedge q''(c),q'(b)\leftarrow q(b')\wedge b=b'$ and $q''(c)\leftarrow q(c')\wedge c=c'$.
  \item We associate each relation symbol~$q$ with a unique vector of pair-wise distinct argument variables~$\bar x_q$, and rewrite every occurrence of $q$ to the form~$q(\bar x_q)$. In addition, all the argument vectors~$\bar x_q$ are kept disjoint.
\end{inparaenum}
%
The normalized CHCs from Figure~\ref{while-C-program-Horn-clauses-simplified} are shown in Table~\ref{tab:data-flow-and-guard-for-horn-clauses}.

\paragraph{Splitting constraints into control- and data-flow formulas.}
We can rewrite the constraint $\varphi$ to a conjunction
$
\varphi=\varphi_{1} \wedge \cdots \wedge \varphi_{k},~k\in \mathbb{N}~.
$
%
The sub-formula~$\varphi_i$ is called a ``data-flow sub-formula'' if and only if it can be rewritten to the form $x=t(\bar y)$ such that
\begin{inparaenum}[(i)]
  \item $x$ is one of the arguments in head $q(\bar x_q)$;
  \item $t(\bar{y})$ is a term over variables $\bar{y}$, where each element of $\bar{y}$ is an argument of some body literal $q'(\bar x_{q'})$.
\end{inparaenum}
We call all other $\varphi_j$ ``control-flow sub-formulas''.
A constraint $\varphi$ can then be represented by
$\varphi=g_{1} \wedge \cdots \wedge g_{m} \wedge d_{1} \wedge \cdots \wedge d_{n}$, where $m,n \in \mathbb{N}$
and $g_{i}$ and $d_{j}$ are the control- and data-flow sub-formulas, respectively.
The control and data flow sub-formulas for the normalized CHCs of our running example are shown in Table~\ref{tab:data-flow-and-guard-for-horn-clauses}.

\begin{table*}[tb]
\caption{Control- and data-flow sub-formula in constraints for the normalized CHCs from Figure~\ref{while-C-program-Horn-clauses-simplified}} \label{tab:data-flow-and-guard-for-horn-clauses}
\begin{tabular}{p{5.3cm}@{~~~~~~~}l@{~~~~~~~}l}
\hline
 Normalized CHCs & Control-flow sub-formula & Data-flow sub-formula \\
\hline
$L(x,y,n)  \leftarrow  n \geq 0 \wedge x = n \wedge y = n$   & $ n \geq 0$ &  $x = n,  y = n$   \\
\hline
$L(x,y,n)  \leftarrow  L'(x',y',n') \wedge x \neq 0 \wedge x = x' -1 \wedge y = y' -1$  & $x \neq 0$ &  $x=x' -1, y=y' -1$   \\
\hline
$L'(x',y',n')  \leftarrow  L(x,y,n) \wedge x' = x \wedge y' = y \wedge n' = n$   & empty &  $x' = x, y' = y, n' = n $  \\
\hline
$\textit{false}  \leftarrow   L(x,y,n) \wedge x = 0 \wedge y \neq 0$  & $y \neq 0$ &  $x = 0$   \\
\hline
\end{tabular}
\end{table*}

\paragraph{Constructing a \HyperedgeHornGraphAbbrev.}
The \HyperedgeHornGraphAbbrev represents program control- and data-flow by guarded control-flow hyperedges $\textit{CFHE}$ and data-flow hyperedges $\textit{DFHE}$.
A $\textit{CFHE}$ edge denotes the flow of control from the body to head literals of the CHC.
A $\textit{DFHE}$ edge denotes how data flows from the body to the head.
Both control- and data-flow are guarded by the control flow sub-formula.

\begin{figure*}[tb]
\centering
  \includegraphics[width=\linewidth]{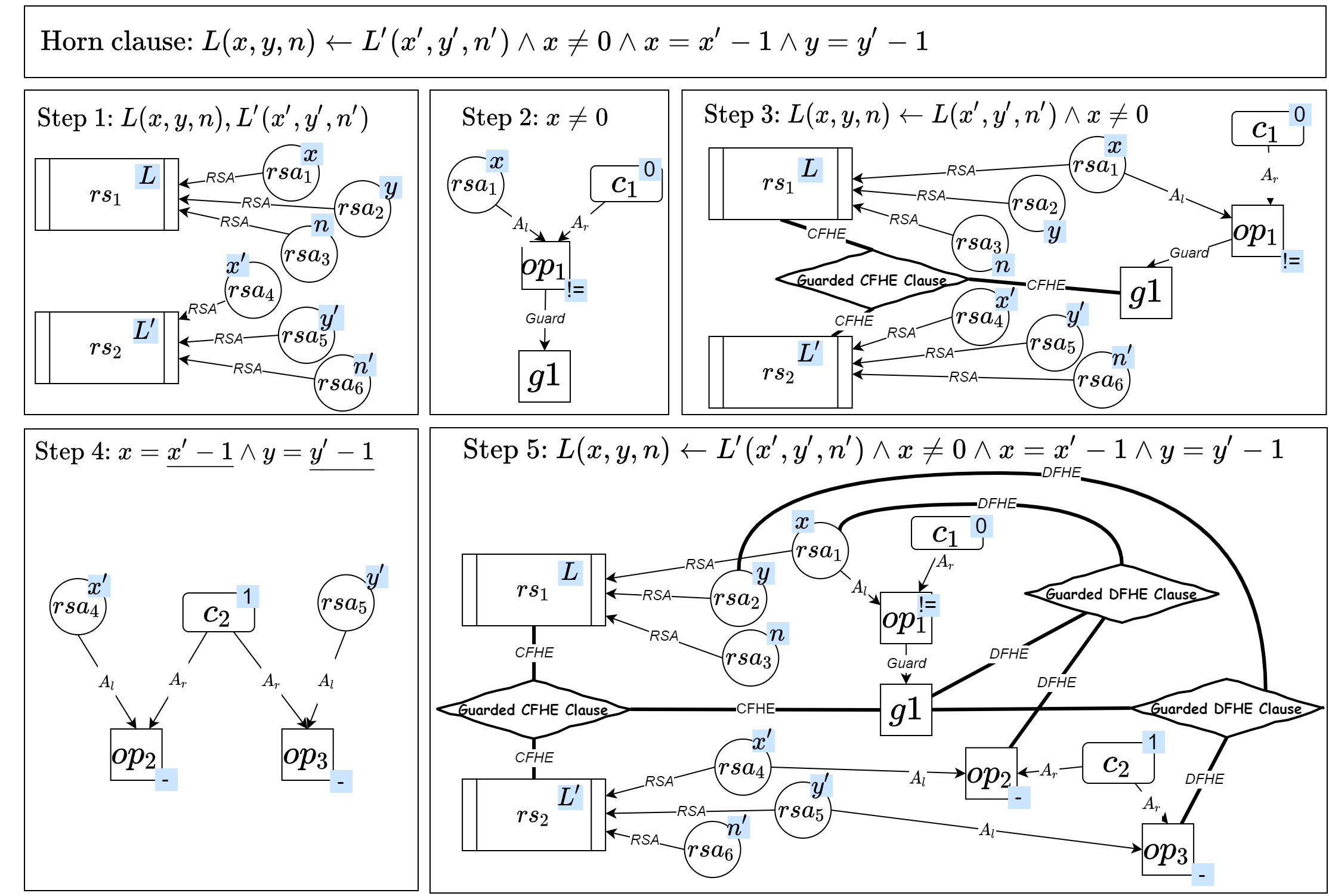}\\
  \caption{Construct the \HyperedgeHornGraphAbbrev from the CHC $L(x,y,n)  \leftarrow  L'(x',y',n') \wedge x \neq 0 \wedge x=x' -1 \wedge y=y' -1$.}\label{hyperedge-graph-step-example}
\end{figure*}

Constructing the \HyperedgeHornGraphAbbrev for a normalized CHC $L(x,y,n)  \leftarrow  L'(x',y',n') \wedge x \neq 0 \wedge x=x' -1 \wedge y=y' -1$ is shown in Figure~\ref{hyperedge-graph-step-example}.
\ifisPAAR
  The corresponding node and edge types are described in Tables~5 and~6 in the full version of this paper \cite{tech-report}.
\else 
  The corresponding node and edge types are described in Tables~\ref{Hyperedge-Horn-graph-node-definition} and~\ref{Hyperedge-Horn-graph-edge-definition}.
\fi

In the first step, we draw relation symbols and their arguments and build the relation between them.
In the second step, we add a $guard$ node and draw ASTs for the control flow sub-formulas.
In the third step, we construct guarded control-flow edges by connecting the relation symbols in the head and body and the $guard$ node, which connects the root of control flow sub-formulas.
In the fourth step, we construct the ASTs for the right-hand side of every data flow sub-formula.
In the fifth step, we construct the guarded data-flow edges by connecting the left- and right-hand sides of the data flow sub-formulas and the $guard$ node. 
Note that the diamond shapes in Figure~\ref{hyperedge-graph-step-example} are not nodes in the graph but are used to visualize our (ternary) hyperedges of types \textit{CFHE} and \textit{DFHE}.
We show it in this way to visualize \HyperedgeHornGraphAbbrev better. 

\ifisPAAR

\else 
\paragraph{Formal definition of \HyperedgeHornGraphAbbrev.}
A \HyperedgeHornGraphAbbrev $HG=(V,\mathit{HE},R^{\HyperedgeHornGraphAbbrev},X^{\HyperedgeHornGraphAbbrev},\ell)$ consists of a set of nodes $v\in V$, a set of typed hyperedge $\mathit{HE} \in V^{*} \times R^{\HyperedgeHornGraphAbbrev}$ where $V^{*}$ is a list of node from $V$, a set of node types $x\in X^{\HyperedgeHornGraphAbbrev}$ (Table~\ref{Hyperedge-Horn-graph-node-definition}),a set of edge types $r\in R^{\HyperedgeHornGraphAbbrev}$ (Table~\ref{Hyperedge-Horn-graph-edge-definition}), and a map $\ell: v\rightarrow x$. 
Here, $(v_{1},v_{2},\cdots,v_{n},r) \in HE$ denotes a hyperedge for a list of nodes $(v_{1},\cdots,v_{n})$ with edge type $r$. The node types are used to generate the initial feature $x_{v}$, a real-valued vector, in \hyperedgeGNN.  

The guarded $\textit{CFHE}$ is a typed hyperedge $(v_{1},\cdots, v_{n},g,\textit{CFHE}) \in \textit{HE}$, where the type of $v_{1},\cdots,v_{n}$ is $rs$. Since \hyperedgeGNN has more stable performance with fixed number of node in one edge type, we transform the hyperedge $(v_{1},\cdots, v_{n},g,\mathit{CFHE}) \in HE$ with variable number of nodes to a set of ternary hyperedges (i.e., $\{(v_{1},v_{2},g,$ $\mathit{CFHE}),(v_{1},v_{3},g,\mathit{CFHE}),\cdots,(v_{1},v_{n},g,\mathit{CFHE})\}$).

The guarded \textit{DFHE} is a typed ternary hyperedge $(v_{i},v_{j},g, \mathit{DFHE}) \in \mathit{HE}$, where $v_{i}$'s type is $rsa$ and $v_{j}$'s type could be one of $\{op,c,v\}$.

\begin{table*}[tb]
\caption{Node types for the \HyperedgeHornGraphAbbrev. Note that Tables~\ref{layer-Horn-graph-node-definition} and~\ref{Hyperedge-Horn-graph-node-definition} use some same node types because they represent the same elements in the CHC. Some abstract nodes, such as \textit{initial} and \textit{guard}, do not have concrete symbol names since they do not directly associate with any element in the CHCs.} \label{Hyperedge-Horn-graph-node-definition}
\begin{tabular}{lp{5cm}lp{1cm}}
\hline
Node types $X^{CDHG}$ & Explanation & Elements in CHCs & Shape\\
\hline
\textit{relation symbol} ($rs$)  & Relation symbols in head or body  &  $L$ & \icon{graph/process}    \\
\textit{initial}  &  Initial state &   $\emptyset$  & \icon{graph/process}\\
\textit{false}  &   \textit{false} state   &   \textit{false}    & \icon{graph/process}      \\
\textit{relation symbol argument} ($rsa$)  & Arguments of the relation symbols &   $x, y$  & \icon{graph/circle} \\
\textit{variables} ($var$)   & Free variables  &   $n$  & \icon{graph/circle}  \\
\textit{operator} ($op$)    &  Operators over a theory &   =, +     & \icon{graph/square}  \\
\textit{constant} ($c$)    &   Constant over a theory  &  0, 1, $true$  & \icon{graph/round-rectangular}\\
\textit{guard} ($g$) & Guard for \textit{CFHE} and \textit{DFHE}  & $\emptyset$ & \icon{graph/square}\\
\hline
\end{tabular}
\end{table*}

\begin{table*}[tb]
\caption{Edge types for the \HyperedgeHornGraphAbbrev.  $rs, rsa, var, op, c, etc.$ are node types from Table~\ref{Hyperedge-Horn-graph-node-definition}.} \label{Hyperedge-Horn-graph-edge-definition}
\begin{tabular}{p{2.8cm}p{1.5cm}p{3.5cm}p{5cm}}
\hline
Edge type $R^{CDHG}$ & Edge arity & Definition & Explanation\\
\hline
Control Flow Hyperedge (\textit{CFHE})  & Ternary  &  $(rs_{1},rs_{2}~|~\textit{false},g,$ $\textit{CFHE})$ & Connects the $rs$ node in body and head, and abstract \textit{guard} node
\\
\hline
Data Flow Hyperedge (\textit{DFHE}) & Ternary &  $(a,op~|~c~|~var,g, \textit{DFHE})$ & Connects the root node of right-hand and left-hand side of data flow sub-formulas, and a abstract \textit{guard} node
\\
\hline
Guard (\textit{Guard})  & Binary  &   $(op~|~c,g,\textit{Guard})$ & Connects all roots of ASTs of control flow sub-formulas and the $guard$ node \\
\hline
Relation Symbol Argument (\textit{RSA})  & Binary & $(a,rs,\textit{RSA})$ & Connects $rsa$ nodes and their $rs$ node \\
\hline
AST\_left ($A_{l}$)   & Binary  & $(op,op~|~var~|~c ,A_{l})$ & Connects left-hand side element of a binary operator or an element from a unary operator \\
\hline
AST\_right ($A_{r}$)   & Binary  & $(op,op~|~var~|~c,A_{r})$ & It connects right-hand side element of a binary operator \\
\hline
\end{tabular}
\end{table*}

\paragraph{An abstract example of \HyperedgeHornGraphAbbrev.}
Except for the concrete example, we give a abstract example to describe how to construct the \HyperedgeHornGraphAbbrev.
After the preprocessings (normalization and splitting the constraint to guard and data flow sub-formulas), the CHC $H \leftarrow B_{1}  \wedge \cdots \wedge B_{n} \wedge \varphi$ in Section~\ref{section-banckground} can be re-written to
\begin{equation}\label{horn-clause-definition-2}
 q_{1}(\bar{t_{1}}) \leftarrow q_{2}(\bar{t_{2}}) \wedge \cdots \wedge q_{k}(\bar{t_{k}}) \wedge g_{1}  \wedge \cdots g_{m} \wedge d_{1} \wedge \cdots d_{n}, (m,n,k \in \mathbb{N}).
\end{equation}

We can construct the corresponding \HyperedgeHornGraphAbbrev using Algorithm~\ref{hyperedge-graph-algorithm}, in which the input is a set of CHC and the output is a \HyperedgeHornGraphAbbrev $HG=(V,\mathit{HE},R^{\HyperedgeHornGraphAbbrev},X^{\HyperedgeHornGraphAbbrev},\ell)$.
The step-wise constructing process for the CHC in~(\ref{horn-clause-definition-2}) is visualized in Figure~\ref{abstract-hyperedge-horn-graph}.

\fi

%% file: sections/4-hyperedge-graph-neural-networks.tex
\section{Relational Hypergraph Neural Network}\label{HyperedgeGraphNeuralNetworks}

\begin{figure}[p]
    \centering
      \includegraphics[width=0.6\linewidth]{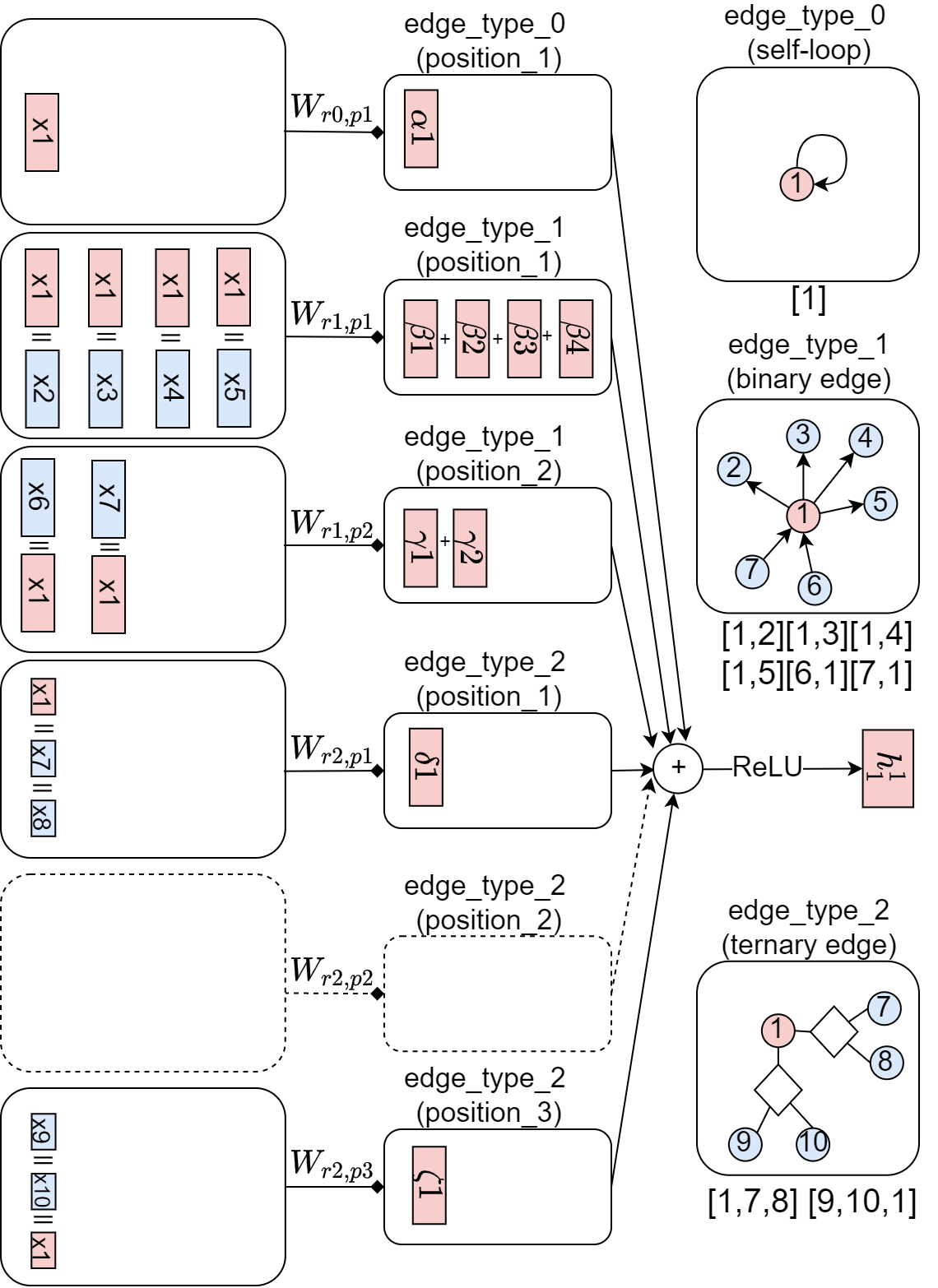}\\
      \caption{An example to illustrate how to update node representation for \hyperedgeGNN using \eqref{Hyperedge-GNN-definition}. At the right-hand side, there are three types of edges connected with node 1. We compute the updated representation $h_{1}^{1}$ for node 1 at the time step 1. $\|$ means concatenation. $x_{i}$ is the initial feature vector of node $i$. The red blocks are the trace of the updating for node 1. The edge type 1 is a unary edge and is a self-loop. It has one set of learnable parameters as the update function i.e., $W_{r0,p1}$. The edge type 2 is binary edge, it has two update functions i.e., $W_{r1,p1}$ and $W_{r1,p2}$. Node 1 is in the first position in edge [1,2], [1,3], [1,4], and [1,5], so the concatenated node representation will be updated by $W_{r1,p1}$ . On the other hand, for the other two edges [6,1] and [7,1], node 1 is in the second position, so the concatenated node representation will be updated by $W_{r1,p2}$. For edge type 3, the same rule applies, i.e., depending on node 1's position in the edge, the concatenated node representation will go through different parameter sets. Since there is no edge that node 1 is in the second position, we use a dashed box and arrow to represent it. The aggregation is to add all updated representations from different edge types. }\label{Hyperedge-GCN}
    \end{figure}

Different from regular graphs, which connect nodes by binary edges, \HyperedgeHornGraphAbbrev includes hyperedges which connect arbitrary number of nodes. Therefore, we extend R-GCN to \hyperedgeGNN to handle hypergraphs.
Concretely, to compute a new representation of a node $v$ at timestep $t$, we consider all hyperedges $e$ that involve $v$.
For each such hyperedge we create a ``message'' by concatenating the representations of \emph{all} nodes involved in that hyperedge and multiplying the result with a learnable matrix $W_{r,p}^t$, where $r$ is the type of the relation and $p$ the position that $v$ takes in the hyperedge.
Intuitively, this means that we have one learnable matrix for each distinguishable way a node can be involved in a relation.
Hence, the updating rule for node representation at time step $t$ in \hyperedgeGNN is
\begin{equation}\label{Hyperedge-GNN-definition}
  h_{v}^{t} = {\rm ReLU}(\sum_{r\in R}\sum_{p\in P_{r}}
  \sum_{e \in E_v^{r,p}}
  W_{r,p}^{t}\cdot \| [ h_{u}^{t-1}\mid u\in e ] ),
\end{equation}
where $\| \{\cdot \}$ denotes concatenation of all elements in a set, $r\in R = \{r_i\mid i\in \mathbb{N} \}$ is the set of edge types (relations), $p\in P_{r}=\{p_{j} \mid j\in \mathbb{N} \}$ is the set of node positions under edge type $r$, $W_{r,p}^{t}$ denotes learnable parameters when the node is in the $p$th position with edge type $r$, and  $e\in E_v^{r,p}$ is the set of hyperedges of type $r$ in the graph in which node $v$ appears in position $p$, where $e$ is a list of nodes.
The updating process for the representation of node $v$ at time step 1 is illustrated in Figure \ref{Hyperedge-GCN}.

Note that different edge types may have the same number of connected nodes. For instance, in \HyperedgeHornGraphAbbrev, $\textit{CFHE}$ and $\textit{DFHE}$ are both ternary edges.

Overall, our definition of \hyperedgeGNN is a generalization of R-GCN. Concretely, it can directly be applied to the special-case of binary graphs, and in that setting is slightly more powerful as each message between nodes is computed using the representations of both source and target nodes, whereas R-GCN only uses the source node representation.

\subsection{Training Model}\label{trainingModel}
The end-to-end model consists of three components: the first component is an embedding layer, which learns to map the node's type (encoded by integers) to the initial feature vectors;
the second component is \hyperedgeGNN, which learns program features; the third component is a set of fully connected neural networks, which learns to solve a specific task by using gathered node representations from \hyperedgeGNNs. All parameters in these three components are trained together. Note that different tasks may gather different node representations. For example, Task~1 gathers all node representations, while Task~2 only gathers node representations with type $rs$.  

We set all vector lengths to 64, i.e., the embedding layer output size, the middle layers' neuron size in \hyperedgeGNNs, and the layer sizes in fully connected neural networks are all 64. The maximum training epoch is 500, and the patient is 100. The number of message passing steps is 8 (i.e., \eqref{Hyperedge-GNN-definition} is applied eight times). For the rest of the parameters (e.g., learning rate, optimizer, dropout rate, etc.), we use the default setting in the tf2\_gnn framework.
We set these parameters empirically according to the graph size and the structure. We apply these fixed parameter settings for all tasks and two graph representations without fine-tuning.

%% file: sections/5-proxy-tasks.tex
\section{Proxy Tasks}
We propose five proxy tasks with increasing difficulty to systematically evaluate the \hyperedgeGNN on the two graph representations. Tasks 1 to 3 evaluate if \hyperedgeGNN can solve general problems in graphs. In contrast, Tasks 4 and 5 evaluate if combining our graph representations and \hyperedgeGNN can learn program features to solve the encoded program verification problems. We first describe the learning target for every task and then explain how to produce training labels and discuss the learning difficulty.

\paragraph{Task 1: \taskOne .}
For both graph representations, the \hyperedgeGNN model performs binary classification on all nodes to predict if the node type is a relation symbol argument ($rsa$) and the metric is accuracy.
The binary training label is obtained by reading explicit node types. This task evaluates if \hyperedgeGNN can differentiate explicit node types.
This task is easy because the graph explicitly includes the node type information in both typed nodes and edges.

\paragraph{Task 2: \taskTwo .}
For both graph representations, the \hyperedgeGNN model performs regression on nodes with type $rs$ to predict how many times the relation symbols occur in all clauses. The metric is mean square error.
The training label is obtained by counting the occurrence of every relation symbol in all clauses. This task is designed to see if \hyperedgeGNN can correctly perform a counting task.
For example, the relation symbol $L$ occurs four times in all CHCs in Figure~\ref{tab:data-flow-and-guard-for-horn-clauses}, so the training label for node $L$ is value 4.
This task is harder than Task 1 since it needs to count the connected binary edges or hyperedges for a particular node.

\paragraph{Task 3: \taskThree .}
For both graph representations, the \hyperedgeGNN model performs binary classification on nodes with type $rs$ to predict if this node is an SCC (i.e., in a cycle) and the metric is accuracy. 
The binary training label is obtained using Tarjan's algorithm~\cite{Tarjan1972DepthFirstSA}.
For example, in Figure~\ref{tab:data-flow-and-guard-for-horn-clauses}, $L$ is an SCC because $L$ and $L'$ construct a cycle by $L \leftarrow L'$ and $L' \leftarrow L$.
This task is designed to evaluate if \hyperedgeGNN can recognize general graph structures such as cycles.
This task requires the model to classify a graph-theoretic object (SCC), which is harder than the previous two tasks since it needs to approximate a concrete algorithm rather than classifying or counting explicit graphical elements.

\paragraph{Task 4: \taskFour .}
For both graph representations, we train two independent \hyperedgeGNN models which perform binary classification on nodes with type $rsa$ to predict if individual arguments have (a)~lower and (b)~upper bounds in the least solution of a set of CHCs, and the metric is accuracy.
To obtain the training label, we apply the Horn solver Eldarica to check the correctness of guessed (and successively increased) lower and upper arguments bounds; arguments for which no bounds can be shown are assumed to be unbounded. We use a timeout of 3~s for the lower and upper bound of a single argument, respectively. The overall timeout for extracting labels from one program is 3~hours.
For example, consider the CHCs in Fig.~\ref{while-C-program-Horn-clauses-simplified}. The CHCs contain a single relation symbol~$L$; all three arguments of $L$ are bounded from below but not from above.
This task is (significantly) harder than the previous three tasks, as boundedness of arguments is an undecidable property that can, in practice, be approximated using static analysis methods.


\paragraph{Task 5: \taskFive}
This task consists of two binary classification tasks on nodes with type $\textit{guard}$ (for \HyperedgeHornGraphAbbrev), and with type $\textit{clause}$ (for \layerHornGraph) to predict if a clause occurs in the counter-examples. Those kinds of nodes are unique representatives of the individual clauses of a problem. The task focuses on unsatisfiable sets of CHCs. Every unsatisfiable clause set gives rise to a set of minimal unsatisfiable subsets (MUSes); MUSes correspond to the minimal CEs of the clause set. Two models are trained independently to predict whether a clause belongs to (a)~the intersection or (b)~the union of the MUSes of a clause set. The metric for this task is accuracy. 
We obtain training data by applying the Horn solver Eldarica~\cite{8603013}, in combination with an optimization library that provides an algorithm to compute MUSes\footnote{\url{https://github.com/uuverifiers/lattice-optimiser/}}.
This task is hard, as it attempts the prediction of an uncomputable binary labelling of the graph.



%
%

%% file: sections/6-evaluation.tex
\section{Evaluation}

We first describe the dataset we use for the training and evaluation and then analyse the experiment results for the five proxy tasks.

\subsection{Benchmarks and Dataset}

Table \ref{benchmark-table} shows the number of labelled graph representations from a collection of CHC-COMP benchmarks~\cite{chcBenchmark}. All graphs were constructed by first running the preprocessor of Eldarica~\cite{8603013} on the clauses, then building the graphs as described in Section~\ref{sec:graphs}, and computing training data.
For instance,  in the first four tasks we constructed 2337 \layerHornGraphsSpace with labels from 8705 benchmarks in the CHC-COMP LIA-Lin track (linear Horn clauses over linear integer arithmetic). The remaining 6368 benchmarks are not included in the learning dataset because when we construct the graphs, (1)~the data generation process timed out, or (2)~the graphs were too big (more than 10,000~nodes), or (3)~there was no clause left after simplification.
In Task 5, since the label is mined from CEs, we first need to identify unsat benchmarks using a Horn solver (1-hour timeout), and then construct graph representations. We obtain 881 and 857 \layerHornGraphsSpace when we form the labels for Task 5 (a) and (b), respectively, in LIA-Lin. 

Finally, to compare the performance of the two graph representations, we align the dataset for both two graph representations to have 5602 labelled graphs for the first four tasks. For Task 5 (a) and (b), we have 1927 and 1914 labelled graphs, respectively. We divide them to train, valid, and test sets with ratios of 60\%, 20\%, and 20\%. We include all corresponding files for the dataset in a Github repository \footnote{\url{https://github.com/ChenchengLiang/Horn-graph-dataset}}. 

\begin{table*}[tb]
\caption{The number of labeled graph representations extracted from a collection of CHC-COMP benchmark ~\cite{chcBenchmark}. For each SMT-LIB file, the graph representations for Task 1,2,3,4 are extracted together using the timeout of 3 hours, and for task 5 is extracted using 20 minutes timeout. Here, T. denotes Task.}
\label{benchmark-table}
\begin{tabular}{l|l|l|l|l|l|l|l|l}
\hline
     & \multicolumn{2}{c|}{SMT-LIB files} & \multicolumn{3}{c|}{\LayerHornGraphs} & \multicolumn{3}{c}{\HyperedgeHornGraphAbbrevs} \\
\hline
                   & Total & Unsat & T. 1-4 & T. 5 (a) & T. 5 (b) & T. 1-4 & T. 5 (a) & T. 5 (b)\\
  \hline
    Linear LIA      & 8705  & 1659 & 2337 & 881  & 857  & 3029 & 880  & 883  \\
 \hline
    Non-linear LIA  & 8425  & 3601 & 3376 & 1141 & 1138 & 4343 & 1497 & 1500  \\
\hline
    Aligned           & 17130 & 5260 & 5602 & 1927 & 1914 & 5602 & 1927 & 1914  \\
\hline
\end{tabular}
\end{table*}

\subsection{Experimental Results for Five Proxy Tasks}

From Table \ref{experiment-results-task-1345}, we can see that for all binary classification tasks, the accuracy for both graph representations is higher than the ratios of the dominant labels. For the regression task, the scattered points are close to the diagonal line. These results show that \hyperedgeGNN can learn the syntactic and semantic information for the tasks rather than performing simple strategies (e.g., fill all likelihood by 0 or 1). Next, we analyse the experimental results for every task.

\paragraph{Task 1: \taskOne.}
When the task is performed in the \layerHornGraph, the  accuracy of prediction is 100\%, which means \hyperedgeGNN can perfectly differentiate if a node is a relation symbol argument $rsa$ node.
When the task is performed in \HyperedgeHornGraphAbbrev, the accuracy is close to 100\% because, unlike in the \layerHornGraph, the number of incoming and outgoing edges are fixed (i.e., $RSA$ and $AI$), the $rsa$ nodes in \HyperedgeHornGraphAbbrev may connect with a various number of edges (including $RSA$, $AST\_1$, $AST\_2$, and $\textit{DFHE}$) which makes \hyperedgeGNN hard to predict the label precisely.

Besides, the data distribution looks very different between the two graph representations because the normalization of CHCs introduces new clauses and arguments. For example, in the simplified CHCs in Figure \ref{while-C-program-Horn-clauses-simplified}, there are three arguments for the relation symbol $L$, while in the normalized clauses in Figure \ref{tab:data-flow-and-guard-for-horn-clauses}, there are six arguments for two relation symbols $L$ and $L'$. If the relation symbols have a large number of arguments, the difference in data distribution between the two graph representations becomes larger. Even though the predicted label in this task cannot directly help solve the CHC-encoded problem, it is important to study the message flows in the \hyperedgeGNNs.

\begin{figure*}[tb]
\begin{minipage}{.48\linewidth}
\includegraphics[width=1\linewidth]{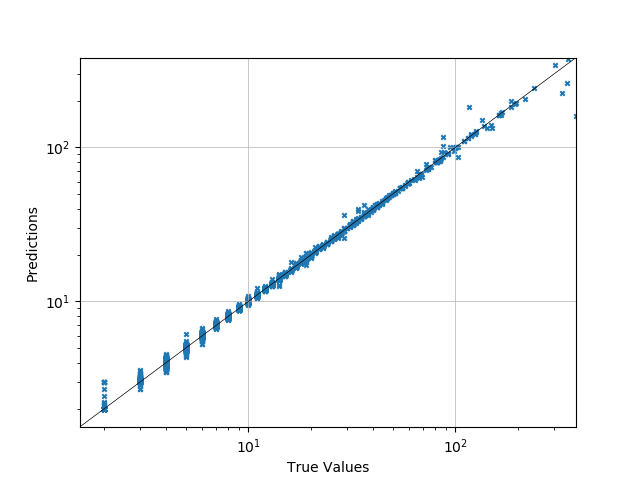}
\subcaption{Scatter plot for \HyperedgeHornGraphAbbrev. The total $rs$ node number is 16858. The mean square error is 4.22.}\label{ScatterPlotTask2Hyperedge}
\end{minipage}
\hfill
\begin{minipage}{.48\linewidth}
\includegraphics[width=1\linewidth]{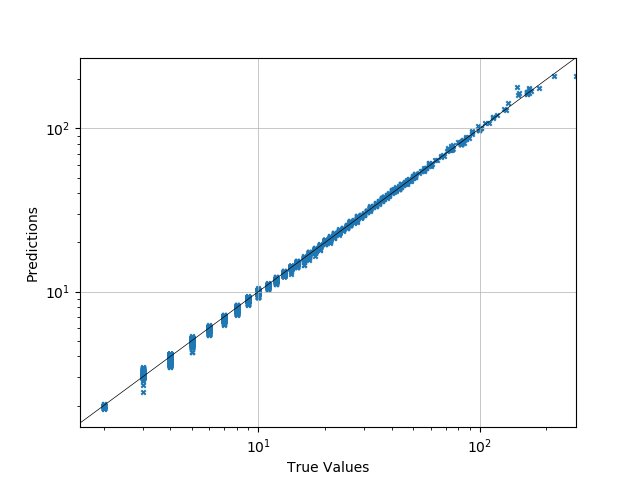}
\subcaption{Scatter plot for \layerHornGraph. The total $rs$ node number is 11131. The mean square error is 1.04.}\label{ScatterPlotTask2layer}
\end{minipage}\caption{Scatter plot for Task 2. The x- and y-axis are in logarithmic scales.}\label{ScatterPlotTask2}
\end{figure*}

\paragraph{Task 2: \taskTwo.}
In the scattered plots in Figure~\ref{ScatterPlotTask2}, the x- and y-axis denote true and the predicted values in the logarithm scale, respectively. The closer scattered points are to the diagonal line, the better performance of predicting the number of relation symbol occurrences in CHCs. Both \HyperedgeHornGraphAbbrevSpace and \layerHornGraphSpace show good performance (i.e., most of the scattered points are near the diagonal lines).
This syntactic information can be obtained by counting the \textit{CFHE} and \textit{RSI} edges for \HyperedgeHornGraphAbbrevSpace and \layerHornGraph, respectively.
When the number of nodes is large, the predicted values are less accurate. We believe this is because graphs with a large number of nodes have a more complex structure, and there is less training data. Moreover, the mean square error for the \HyperedgeHornGraphAbbrevSpace is larger than the \layerHornGraph because normalization increases the number of nodes and the maximum counting of relation symbols for \HyperedgeHornGraphAbbrev, and the larger range of the value is, the more difficult for regression task. Notice that the number of test data (1115) for this task is less than the data in the test set (1121) shown in Table~\ref{benchmark-table} because the remaining six graphs do not have a $rs$ node.

\paragraph{Task 3: \taskThree.}
The predicted high accuracy for both graph representations shows that our framework can approximate Tarjan's algorithm~\cite{Tarjan1972DepthFirstSA}. In contrast to Task 2, even if the \HyperedgeHornGraphAbbrevSpace has more nodes than the \layerHornGraphSpace on average, the \HyperedgeHornGraphAbbrevSpace has better performance than the \layerHornGraphSpace, which means the control and data flow in \HyperedgeHornGraphAbbrevSpace can help \hyperedgeGNN to learn graph structures better. For the same reason as task 2, the number of test data (1115) for this task is less than the data in the test set (1121).

\paragraph{Task 4: \taskFour.}
For both graph representations, the accuracy is much higher than the ratio of the dominant label. 
Our framework can predict the answer for undecidable problems with high accuracy, which shows the potential for guiding CHC solvers. The \HyperedgeHornGraphAbbrevSpace has better performance than the \layerHornGraph, which might be because predicting argument bounds relies on semantic information. The number of test data (1028) for this task is less than the data in the test set (1121) because the remaining 93 graphs do not have a $rsa$ node.

\paragraph{Task 5: \taskFive.}
For Task~(a) and (b), the overall accuracy for two graph representations is high. We manually analysed some predicted results by visualizing the (small) graphs\footnote{\url{https://github.com/ChenchengLiang/Horn-graph-dataset/tree/main/example-analysis/task5-small-graphs}}. We identify some simple patterns that are learned by \hyperedgeGNNs. For instance, the predicted likelihoods are always high for the $rs$ nodes connected to the $\mathit{false}$ nodes. 
One promising result is that the model can predict all labels perfectly for some big graphs\footnote{\url{https://github.com/ChenchengLiang/Horn-graph-dataset/tree/main/example-analysis/task5-big-graphs}} that contain more than 290 clauses, which confirms that the \hyperedgeGNN is learning certain intricate patterns rather than simple patterns.
In addition, the \HyperedgeHornGraphAbbrevSpace has better performance than the \layerHornGraphSpace, possibly because semantic information is more important for solving this task.


\begin{table*}[tb]
\caption{Experiment results for Tasks 1,3,4,5. Both the fourth and fifth tasks consist of two independent binary classification tasks. Here, $+$ and $-$ stands for the positive and negative label. The T and P represent the true and predicted labels. The Acc. is the accuracy of binary classifications. The Dom. is dominant label ratio.
Notice that even if the two graph representations originate from the same original CHCs, the label distributions are different since the \HyperedgeHornGraphAbbrev is constructed from normalized CHCs.} \label{experiment-results-task-1345}
\begin{tabular}{l|l|l|l|l|l|l|l|l|l|l}
	\hline
	 \multicolumn{3}{c|}{ } & \multicolumn{4}{c|}{\LayerHornGraphsAbbrev} & \multicolumn{4}{c}{\HyperedgeHornGraphAbbrev}  \\
	 \hline
	Task & Files & \backslashbox{T}{P} & + & - & Acc.& Dom.  & + & - & Acc.  & Dom. \\
	\hline
	\multirow{2}{*}{1} & \multirow{2}{*}{1121} & + & 93863 & 0 & \multirow{2}{*}{100\%} & \multirow{2}{*}{95.1\%} & 142598 & 0 & \multirow{2}{*}{99.9\%} & \multirow{2}{*}{72.8\%}   \\
	\cline{3-5} \cline{8-9}
	 					& & - & 0     & 1835971 & &  & 10 & 381445 & & \\
	\hline
	\multirow{2}{*}{3} & \multirow{2}{*}{1115} & + & 3204 & 133 & \multirow{2}{*}{96.1\%} & \multirow{2}{*}{70.1\%} & 8262 & 58 & \multirow{2}{*}{99.6\%} & \multirow{2}{*}{50.7\%}  \\
	\cline{3-5} \cline{8-9}
							& & - & 301 & 7493 & & & 15 & 8523 & & \\
	\hline
	\multirow{2}{*}{4 (a)} & \multirow{4}{*}{1028} & + & 13685 & 5264 & \multirow{2}{*}{91.2\%} & \multirow{2}{*}{79.7\%}  & 30845 & 4557 & \multirow{2}{*}{94.3\%} & \multirow{2}{*}{75.2\%}  \\
	\cline{3-5} \cline{8-9}
	 & & - & 2928 & 71986 & & & 3566 & 103630 & & \\
	\cline{1-1}\cline{3-11}
	\multirow{2}{*}{4 (b)} & & + & 18888 & 4792 & \multirow{2}{*}{91.4\%} & \multirow{2}{*}{74.8\%} & 41539 & 4360 & \multirow{2}{*}{94.3\%} & \multirow{2}{*}{67.8\%}   \\
	\cline{3-5}\cline{8-9}
	 & & - & 3291 & 66892 & & & 3715 & 92984 & & \\
	\hline
	\multirow{2}{*}{5 (a)} & \multirow{2}{*}{386} & + & 1048 & 281 & \multirow{2}{*}{95.0\%} & \multirow{2}{*}{84.7\%}  & 1230 & 206 & \multirow{2}{*}{96.9\%} & \multirow{2}{*}{86.4\%} \\
	\cline{3-5} \cline{8-9}
	 & & - & 154 & 7163 & & & 121 & 9036 & &\\
	\hline
	\multirow{2}{*}{5 (b)} & \multirow{2}{*}{383} & + & 3030 & 558 & \multirow{2}{*}{84.6\%} & \multirow{2}{*}{53.1\%} & 3383 & 481 & \multirow{2}{*}{90.6\%} & \multirow{2}{*}{54.8\%} \\
	\cline{3-5} \cline{8-9}
	 & & - & 622 & 3428 & & & 323 & 4361 & &\\
	\hline
\end{tabular}
\end{table*}

%% file: sections/8-related-works.tex
\section{Related Work} \label{section-related-work}
%

\paragraph{Learning to represent programs.}

Contextual embedding methods (e.g. transformer~\cite{vaswani2017attention}, BERT~\cite{DBLP:journals/corr/abs-1810-04805}, GPT~\cite{Radford2018ImprovingLU}, etc.) achieved impressive results in understanding natural languages. Some methods are adapted to explore source code understanding in text format (e.g. CodeBERT~\cite{DBLP:journals/corr/abs-2002-08155}, cuBERT~\cite{DBLP:journals/corr/abs-2001-00059}, etc.). But, the programming language usually contains rich, explicit, and complicated structural information, and the problem sets (learning targets) of it~\cite{DBLP:journals/corr/abs-1909-09436,DBLP:journals/corr/abs-2102-04664} are different from natural languages. Therefore, the way of representing the programs and learning models should adapt to the programs' characteristics.
Recently, the frameworks consisting of structural program representations (graph or AST) and graph or tree-based deep learning models made good progress in solving program language-related problems.
For example, 
\cite{DBLP:journals/corr/AllamanisPS16} represents the program by a sequence of code subtokens and predicts source code snippets summarization by a novel convolutional attention network.
Code2vec~\cite{Alon:2019:CLD:3302515.3290353} learns the program from the paths in its AST and predicts semantic properties for the program using a path-based attention model.
\cite{DBLP:journals/corr/MouLJZW14} use AST to represent the program and classify programs according to functionality using the tree-based convolutional neural network (TBCNN).
Some studies focus on defining efficient program representations, others focus on introducing novel learning structures, while we do both of them (i.e. represent the CHC-encoded programs by two graph representations and propose a novel GNN structure to learn the graph representations).

\paragraph{Deep learning for logic formulas.}
Since deep learning is introduced to learn the features from logic formulas,
an increasing number of studies have begun to explore graph representations for logic formulas and corresponding learning frameworks because logic formulas are also highly structured like program languages.
For instance, 
DeepMath~\cite{NIPS2016_6280} had an early attempt to use text-level learning on logic formulas to guide the formal method's search process, in which neural sequence models are used to select premises for automated theorem prover (ATP). 
Later on, FormulaNet~\cite{NIPS2017_6871} used an edge-order-preserving embedding method to capture the structural information of higher-order logic (HOL) formulas represented in a graph format.
As an extension of FormulaNet, \cite{DBLP:journals/corr/abs-1905-10006} construct syntax trees of HOL formulas as structural inputs and use message-passing GNNs to learn features of HOL to guide theorem proving by predicting tactics and tactic arguments at every step of the proof.
LERNA~\cite{inbook-Rawson} uses convolutional neural networks (CNNs)~\cite{NIPS2012_c399862d} to learn previous proof search attempts (logic formulas) represented by graphs to guide the current proof search for ATP.
NeuroSAT~\cite{DBLP:journals/corr/abs-1903-04671,DBLP:journals/corr/abs-1802-03685} reads SAT queries (logic formulas) as graphs and learns the features using different graph embedding strategies (e.g. message passing GNNs)~\cite{4700287,li2017gated,DBLP:journals/corr/GilmerSRVD17}) to directly predict the satisfiability or guide the SAT solver.
Following this trend, we introduce \hyperedgeGNN to learn the program features from the graph representation of CHCs.


\paragraph{Graph neural networks.}
Message passing GNNs~\cite{DBLP:journals/corr/GilmerSRVD17}, such as graph convolutional network (GCN)~\cite{schlichtkrull2017modeling}, graph attention network (GAT)~\cite{2017arXiv171010903V}, and gated graph neural network (GGNN)~\cite{li2017gated} have been applied in several domains ranging from predicting molecule properties to learning logic formula representations. However, these frameworks only apply to graphs with binary edges. Some spectral methods have been proposed to deal with the hypergraph~\cite{inproceedings-GuiHuan,DBLP:journals/corr/abs-1711-10146}. For instance, the hypergraph neural network (HGNN)~\cite{DBLP:journals/corr/abs-1809-09401} extends GCN proposed by~\cite{DBLP:journals/corr/KipfW16} to handle hyperedges. The authors in \cite{bai2020hypergraph} integrate graph attention mechanism~\cite{2017arXiv171010903V} to hypergraph convolution~\cite{DBLP:journals/corr/KipfW16} to further improve the performance.
But, they cannot be directly applied to the spatial domain. Similar to the fixed arity predicates strategy described in LambdaNet~\cite{wei20206lambdanet}, \hyperedgeGNN concatenates node representations connected by the hyperedge and updates the representation depending on the node's position in the hyperedge.

%

%% file: sections/9-conclusion-and-future-works.tex
\section{Conclusion and Future Work}
In this work, we systematically explore learning program features from CHCs using \hyperedgeGNN, using two tailor-made graph representations of CHCs. We use five proxy tasks to evaluate the framework. The experimental results indicate that our framework has the potential to guide CHC solvers analysing Horn clauses. 
In future work, among others we plan to use this framework to filter predicates in Horn solvers applying the CEGAR model checking algorithm. 

%% file: sections/abstract-examples-and-algorithms.tex
     
\begin{figure*}[p]
\centering
  \includegraphics[width=\linewidth]{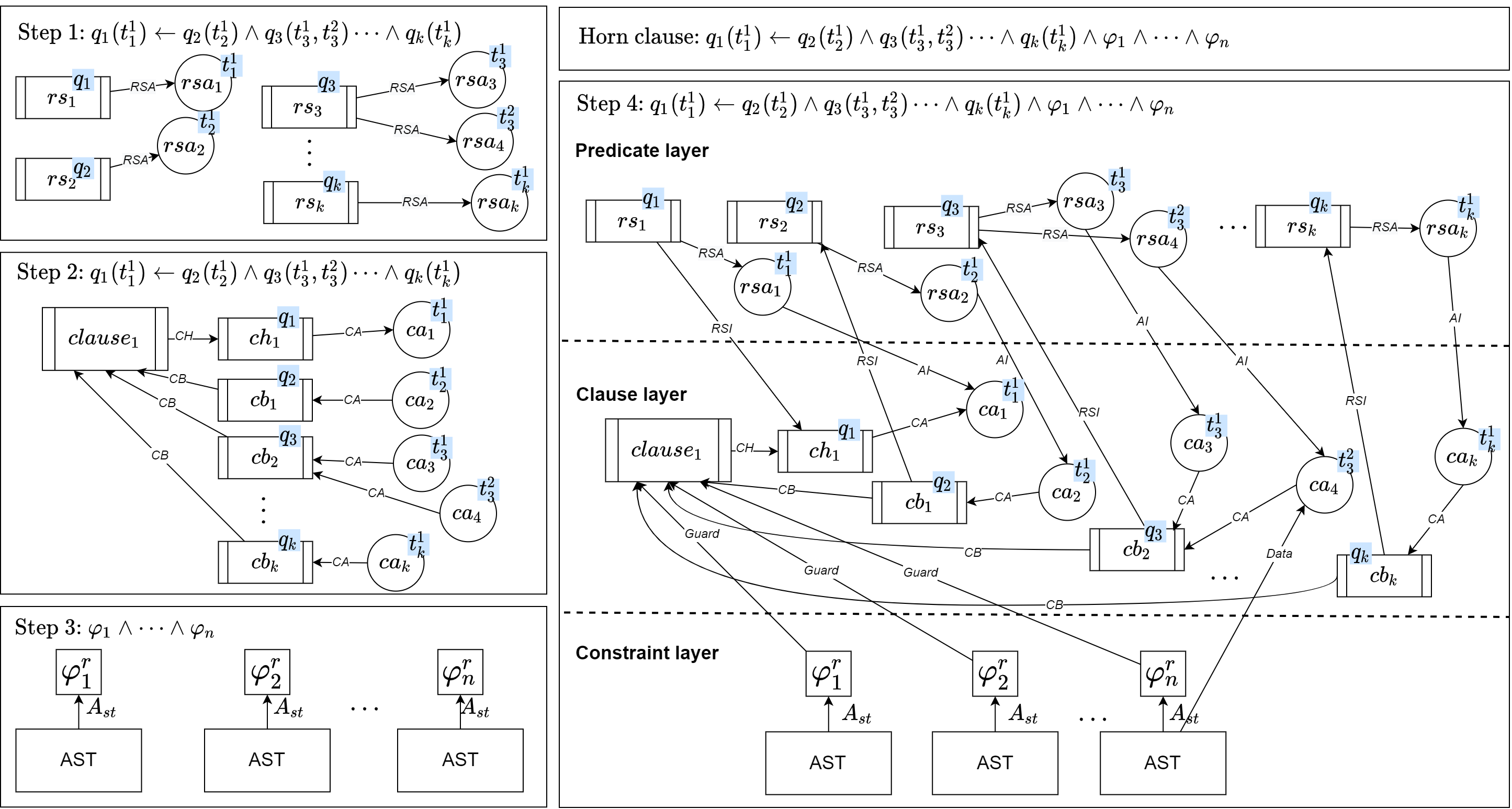}\\
  \caption{An example to illustrate how to construct the \LayerHornGraph from the CHC in (\ref{horn-clause-definition-2}). Here, $\varphi_{i}^{r}$ is the AST root node $\varphi_{i}$.}\label{abstract-layer-horn-graph}
\end{figure*}

\begin{algorithm*}[tb]
\small
\SetKwInput{KwData}{Input}
\SetKwInput{KwResult}{Output}
\caption{Construct a \layerHornGraphSpace from CHCs}\label{layer-graph-algorithm}
\SetAlgoLined
\KwData{ Simplified CHCs $\Pi$}
\KwResult{A \layerHornGraph $\textit{CG}=(V,\textit{BE},R^{CG},X^{CG},\ell)$}
Initialise graph: $V,\textit{BE},\ell=\emptyset,\emptyset,\emptyset$\;

\For{ each clause $C$ in $\Pi$}{
    //step 1: construct the predicate layer\;
    \For{ each relation symbol $q(\bar{t})$ in $C$ }{
        \If{$v_{q} \notin V$}{
            //construct a $rs$ node\;
            add a node $v_{q}$ with type $rs$ to $V$. Update map $\ell$; 
            //construct $rsa$ nodes and connect them to the $rs$ node\;
            \For{ each argument $t$ in $\bar{t}$}{
                add a node $v_{q}^{t}$ with type $rsa$ to $V$. Update map $\ell$\;
                add a edge from $v_{q}$ to $v_{q}^{t}$ to with type $\textit{RSA}$ to $\textit{BE}$ ;\
            }
        }
    }
    //step 2: construct the clause layer\;
    add a node $v_{clause}$ with type $clause$ to $V$. Update map $\ell$;\ 
    add a clause head node $v_{q}^{h}$ with type $ch$ to $V$ for the relation symbol in $C$' head. Update map $\ell$\;
    add a edge from $v_{clause}$ to $v_{q}^{h}$ with type $\textit{CH}$ to $\textit{BE}$ \;
    \For{ each argument $t$ in $\bar{t}$ }{
        add a node $v_{q}^{h,t}$ with type $ca$ to $V$. Update map $\ell$\;
        add a edge from $v_{q}^{h}$ to $v_{q}^{h,t}$ with type $\textit{CA}$ to $\textit{BE}$ \;
    }
    \For{ each relation symbol $q(\bar{t})$ in $C$' body }{
        add a clause body node $v_{q}^{b}$ with type $cb$ to $V$. Update map $\ell$ \;
        add a edge from $v_{q}^{b}$ to $v_{clause}$ to with type $\textit{CB}$ to $\textit{BE}$ \;
        \For{ each argument $t$ in $\bar{t}$ }{
            add a node $v_{q}^{b,t}$ with type $ca$ to $V$. Update map $\ell$\;
            add a edge from $v_{q}^{b,t}$ to $v_{q}^{t}$ to with type $\textit{CA}$ to $\textit{BE}$ \;
        }
    }
    //step 3: construct the constraint layer\;
    \For{ each sub-expression $\phi_{i}$ in the constraint}{
        \For{ each sub-expressions $se$ in $\phi_{i}$ }{
            \If{$v_{se} \notin V$}{
                add a node $v_{se}$ with type $rsa,var,c$ or $op$ to $V$. Update map $\ell$\;
            }
            add edge from $v_{se}$ to the left and right child of $se$ to $\textit{BE}$ with type $A_{st}$\;
        }
    }
    //step 4: add connection between three layers\;
    \For{ each relation symbol node $v_{q}$ in predicate layer with type $rs$ }{
        add edge $(v_{q},v_{q}^{h})$ or $(v_{q}^{b},v_{q})$ to $\mathit{BE}$ with type $RSI$\;
    }
    \For{ each argument node $v_{q}^{t}$ in predicate layer with type $rsa$ }{
        add edge $(v_{q}^{t},v_{q}^{h,t})$ or $(v_{q}^{b,t},v_{q}^{t})$ to $\textit{BE}$ with type $AI$\;
    }
    \For{ each root node of sub-expression $v_{\phi_{i}^{r}}$ }{
        add edge $(v_{\phi_{i}^{r}},v_{clause})$ to $\textit{BE}$ with type Guard\;
    }
    \For{ each node $v_{se}$ in AST tree with type $v$; \text{if} $v_{se}$ is an argument node }{
        add edge $(v_{se},v_{q}^{h,t})$ or $(v_{q}^{b,t},v_{se})$ to $\textit{BE}$ with type Data\;
    }
}
\Return{$V,\textit{HE},\ell$}
\end{algorithm*}

\begin{figure*}[p]
\centering
  \includegraphics[width=\linewidth]{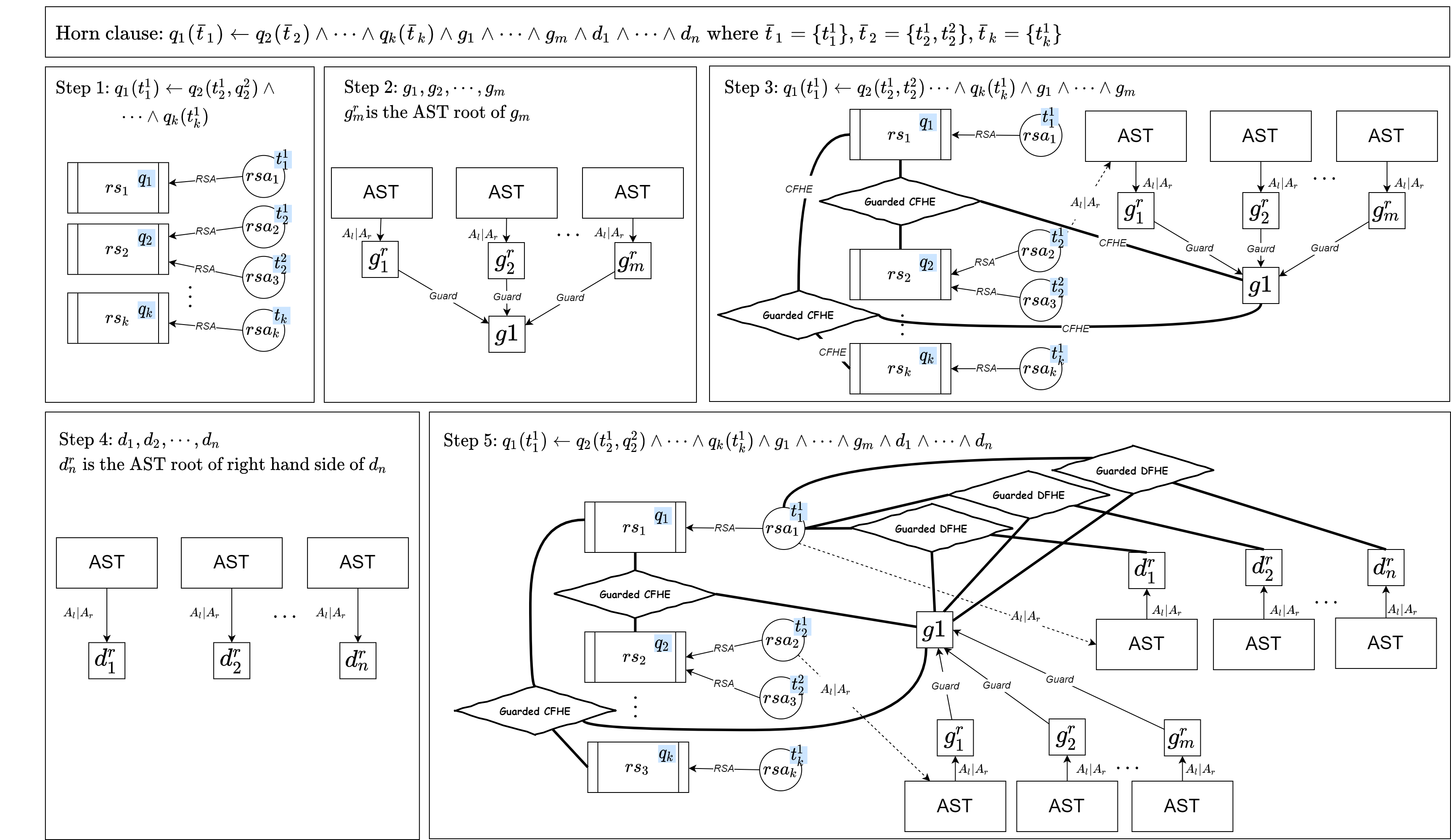}\\
  \caption{An example to illustrate how to construct the \HyperedgeHornGraphAbbrev from the CHC in (\ref{horn-clause-definition-2}). Here, $g_{i}^{r}$ and $d_{j}^{r}$ are the AST root node of control flow sub-formula $g_{i}$ and the right-hand side of the data flow sub-formula, respectively. The ``$|$" connected edge type $A_{l}~|~A_{r}$ means the edge could be type $A_{l}$ or $A_{r}$. }\label{abstract-hyperedge-horn-graph}
\end{figure*}

\begin{algorithm*}[tb]
\small
\SetKwInput{KwData}{Input}
\SetKwInput{KwResult}{Output}
\caption{Construct a \HyperedgeHornGraphAbbrev from CHCs}\label{hyperedge-graph-algorithm}
\SetAlgoLined
\KwData{ Normalized CHCs $\Pi$}
\KwResult{A \HyperedgeHornGraphAbbrevSpace $\mathit{HG}=(V,\textit{HE},R^{\HyperedgeHornGraphAbbrev},X^{\HyperedgeHornGraphAbbrev},\ell)$}
Initialise graph: $V,\textit{HE},\ell=\emptyset,\emptyset,\emptyset$\;

\For{ each clause $C$ in $\Pi$}{
     Add node $v_{initial}$ and $v_{\mathit{false}}$ with type $\textit{initial}$ and $\mathit{false}$ to $V$. Update map $\ell$\;
    //step 1: construct relation symbols and their arguments as typed nodes and add the $RSA$ edge between them\;
    \For{ each relation symbol $q(\bar{t})$ in $C$ }{
        \If{$v_{q} \notin V$}{
            add a node $v_{q}$ with type $rs$ to $V$. Update map $\ell$\;
            \For{ each argument $t$ in $\bar{t}$ }{
                add a node $v_{q}^{t}$ with type $rsa$ to $V$. Update map $\ell$\;
                add a edge from $v_{q}$ to $v_{q}^{t}$ to with type $\textit{RSA}$ to $\textit{HE}$ ;\
            }
        }
    }
    //step 2: construct ASTs for control flow sub-formulas in the constraints and connects the roots of ASTs to an abstract node with type Guard\;
     add a node $v_{g}$ with type $guard$ to $V$. Update map $\ell$;\
    \For{ each $g_{i}$ }{
        \For{ each sub-expression $se$ in $g_{i}$ }{
            \If{$v_{se} \notin V$}{
                add a node $v_{se}$ with type $rsa,var,c,$ or $op$ to $V$. Update map $\ell$\;
            }
            add edge from $v_{se}$ to left and right child of $se$ to $\mathit{HE}$ with type $A_{l}$ and $A_{r}$, respectively\;
        }
        add edge from $g_{i}^{r}$ to $v_{g}$ with type Guard to $\mathit{HE}$, where $g_{i}^{r}$ is the root node of $g_{i}$'s AST\;
    }
    //step 3: construct $\textit{CFHEs}$;\\
    \eIf{$C$'s body $\neq \emptyset$}{
        \For{each relation symbol $q(\bar{t})$ in $C$'s body}{
            add a ternary hyperedge $(v_{q(head)},v_{q(body_{i})},v_{g})$ with type $\mathit{CFHE}$, where $v_{q(head)}$ is the node with type $rs$ or $\mathit{false}$ in the head, and $v_{q(body_{i})}$ is the node with type $rs$ in the body;}
    }{
         add a ternary hyperedge $(v_{q(head)},v_{initial},v_{g})$ with type $\textit{CFHE}$ to $\mathit{HE}$;\
    }

    \For{ each $d_{j}$ }{
        //step 4: construct AST for right-hand side of data flow sub-formula $d_{j}$ \;
        \For{ each sub-expression $se$ in right-hand side of $d_{j}$ }{
            \If{$v_{se} \notin V$}{
                add a node $v_{se}$ with type $rsa,var,c,$ or $op$ to $V$. Update map $\ell$\;
            }
            add edge from $v_{se}$ to left and right child of $se$ to $\textit{HE}$ with type $A_{l}$ or $A_{r}$ \;
        }
        //step 5: construct $\textit{DFHEs}$\;
        add a ternary hyperedge $(v_{q}^{t},d_{j}^{r},v_{g})$ with type $\textit{DFHE}$ to $\textit{HE}$, where $v_{q}^{t}$ is the left-hand side element of $d_{j}$ (a node with type $rsa$) and $d_{i}^{r}$ is the root node of $d_{i}$'s AST (a node with type $rsa,var,c$ or $op$ )\;
    }
}
\Return{$V,\textit{HE},\ell$}
\end{algorithm*}